\documentclass{article}
\usepackage{spconf,amsmath,graphicx}
\usepackage{widetable}
\usepackage{colortbl}
\usepackage{xcolor}
\usepackage{booktabs}
\usepackage{amssymb}
\usepackage{graphicx}
\usepackage{multirow}
\usepackage{widetable}
\usepackage{bm,cite}

\title{MARS-CLIP: Multi-Resolution and Attention Refined Zero-Shot Image Segmentation}
\name{Nagito Saito, Shintaro Ito, Koichi Ito, and Takafumi Aoki}
\address{Graduate School of Information Sciences, Tohoku University, Japan.}
\begin{document}
\ninept
\maketitle
\begin{abstract}
  Contrastive Language-Image Pre-training (CLIP) has demonstrated impressive capabilities in zero-shot transfer but often struggles with dense prediction tasks due to low spatial resolution and the loss of structural information.
  To address these limitations, we propose MARS-CLIP (Multi-resolution and Attention Refined Segmentation for CLIP), a novel framework for zero-shot semantic segmentation.
  Our approach introduces two key strategies: (i) a multi-resolution feature extraction module that fuses local fine-grained features with global context to overcome input resolution constraints, and (ii) an attention refinement mechanism that injects spatial and color biases from intermediate layers into the final self-attention block to accurately restore object boundaries.
  A set of experiments on six public datasets demonstrates that MARS-CLIP significantly outperforms state-of-the-art methods.
\end{abstract}
\section{Introduction}
\label{sec:intro}

Image segmentation, which performs pixel-level object identification and classification, serves as a foundational technology for autonomous driving \cite{Minaee-TPAMI-2022}, medical image analysis \cite{Xu-BE-2024,Xia-NPJPO-2024}, and quality control \cite{Cumbajin-JI-2023,Wu-SSCE-2024}.
In recent years, deep learning-based methods \cite{Long-CVPR-2015,Chen-ECCV-2018,Xie-NeurIPS-2021} have become the major approach due to their robustness and high accuracy; however, they require large-scale datasets with dense pixel-level annotations.
Consequently, extending these methods to recognize unseen classes beyond the training data requires additional annotation and model retraining, resulting in prohibitive time and labor costs.

To address this challenge, zero-shot image segmentation methods \cite{Zhou-ECCV-2022,Li-PR-2025,Bousselham-CVPR-2024}, which can recognize unseen classes based on text descriptions, have attracted significant attention.
Most of these methods leverage Contrastive Language-Image Pre-Training (CLIP) \cite{Radford-ICML-2021} to achieve segmentation by exploiting the similarity between image and text in a shared feature space.
However, CLIP is inherently designed to capture global representations of an entire image and is not optimized for preserving local, pixel-level details.
Therefore, applying CLIP to dense prediction tasks presents two fundamental challenges that remain to be solved.
The first challenge is the insufficient spatial resolution.
Since the feature maps output by the CLIP image encoder are of a fixed low resolution, existing methods such as MaskCLIP \cite{Zhou-ECCV-2022} and SCLIP \cite{Wang-ECCV-2024} often fail to detect tiny objects or accurately delineate complex boundaries.
The second challenge is the loss of spatial structural information.
Although SCLIP \cite{Wang-ECCV-2024} improved spatial consistency by introducing a self-attention mechanism based on Key-Key similarity in the final layer, semantic abstraction in deep ViT layers inevitably dilutes the original shape information.
Spatial layouts preserved in intermediate layers and low-level features like edges and colors are not fully utilized, often leading to ambiguous segmentation results.

In this paper, we propose {\it MARS-CLIP} (Multi-resolution and Attention Refined Segmentation for CLIP), a zero-shot segmentation framework capable of accurately recognizing fine image details through high-definition feature extraction by multi-resolution inputs and attention map refinement using low-level features.
The main contributions of this work are summarized as follows:
First, we introduce a multi-resolution strategy.
By partitioning the input image into local grid patches for encoding, we circumvent input resolution limitations.
These detailed local features are then adaptively fused with global features extracted from the entire image to obtain high-resolution feature maps that preserve context.
Second, we propose a refined attention mechanism by bias injection.
In contrast to the standard Key-Key correlation used in SCLIP \cite{Wang-ECCV-2024}, we inject spatial biases from intermediate layers and color affinity biases, which capture sharp object boundaries, into the attention mechanism.
Furthermore, by eliminating residual connections and feed-forward networks in the final layer, we reduce noise and achieve a balance between semantic consistency and spatial fidelity.
We demonstrate the effectiveness of our proposed method through a set of experiments on PASCAL VOC 2012 \cite{Everingham-IJCV-2015}, PASCAL Context \cite{Mottaghi-CVPR-2014}, ADE20K \cite{Zhou-IJCV-2019}, Cityscapes \cite{Cordts-CVPR-2016}, COCO-Object \cite{Lin-ECCV-2014}, and COCO-Stuff \cite{Caesar-CVPR-2018}, showing significant improvements over state-of-the-art methods.

\begin{figure*}[t]
  \centering
  \includegraphics[width=.9\linewidth]{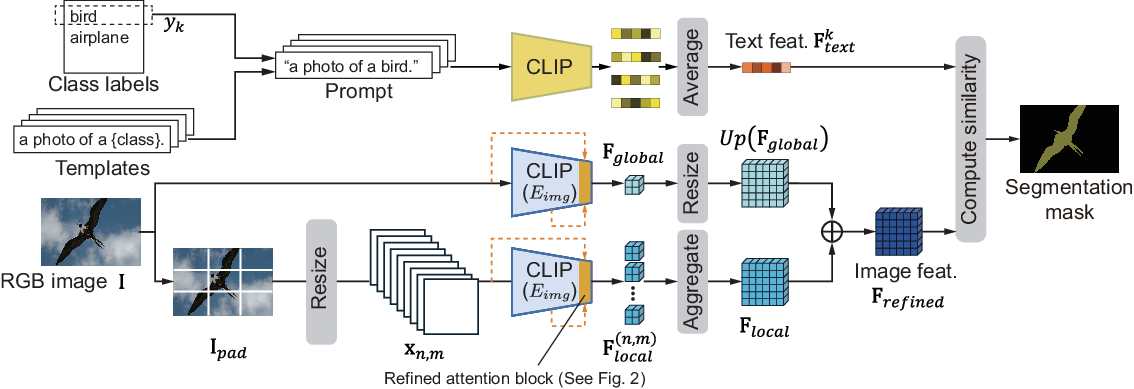} 
  \caption{Overview of MARS-CLIP.
  Global and local features are fused by a multi-resolution strategy.
  The final layer (highlighted in orange) injects spatial and color biases to refine boundaries, as detailed in Fig. \ref{fig:last_transformer}.} 
  \label{fig:overview} 
\end{figure*}

\section{Related Work}
\label{sec:related work}

In this section, we give a brief overview of zero-shot image segmentation utilizing CLIP, focusing on recent approaches centered on attention mechanism improvements and spatial structure preservation.


\subsection{Contrastive Language-Image Pre-training (CLIP)}

CLIP \cite{Radford-ICML-2021} is a vision-language model trained on 400 million image-text pairs, achieving high generalization performance by embedding images and texts into a common feature space.
While CLIP demonstrates excellent capabilities in image-level zero-shot classification, its training process focuses on global feature alignment, making it difficult to apply directly to dense prediction tasks like image segmentation at the pixel level.


\subsection{Zero-Shot Semantic Segmentation}

Early zero-shot segmentation methods \cite{Bucher-NIPS-2019, Baek-ICCV-2021} had limited ability to handle unseen classes due to insufficient general alignment between visual and textual features.
Since the advent of CLIP, research applying its strong zero-shot capabilities to segmentation has accelerated.
A representative method, MaskCLIP \cite{Zhou-ECCV-2022}, achieved segmentation by modifying the last attention layer to extract pixel-wise features, but the resulting prediction masks contained significant noise.
To improve pixel-level feature correspondence, approaches modifying CLIP's attention mechanism itself have become mainstream.
CLIP Surgery \cite{Li-PR-2025} pointed out that standard Query-Key attention causes spatial redundancy and improved feature distinctiveness by introducing Value-Value attention.
SCLIP \cite{Wang-ECCV-2024} revisited the self-attention mechanism and proposed correlative self-attention based on Query-Query and Key-Key similarities, resolving spatial feature entanglement.
Furthermore, ClearCLIP \cite{Lan-ECCV-2024-Clear} identified residual connections in the final ViT layer as a source of segmentation noise and successfully sharpened masks by removing them.
ResCLIP \cite{Yang-CVPR-2025} aims for further accuracy improvement by introducing a residual attention mechanism.
While these methods improve semantic consistency by manipulating CLIP's internal attention mechanisms, the constraint of fixed input resolution remains, leaving challenges in recognizing fine structures and complex boundaries.
In addition to attention mechanism improvements, attempts to reinforce spatial consistency have also been made.
NACLIP \cite{Hajimiri-WACV-2025} imposes smoothing constraints on attention maps between neighboring tokens, while OPMapper \cite{Wang-NIPS-2025} reduces local ambiguity by integrating global context information.
PnP-OVSS \cite{Luo-CVPR-2024} adopts a method of refining masks through an iterative inference process.
Meanwhile, approaches utilizing external foundation models have also been proposed.
ProxyCLIP \cite{Lan-ECCV-2024-Proxy} utilizes DINO's local features, and FreeDA \cite{Barsellotti-CVPR-2024} leverages Stable Diffusion's generative capabilities to expand feature representation.
Although these methods using external knowledge show high performance, they require running multiple models in parallel, making increased computational cost unavoidable.
In this paper, we solve the conventional issues of insufficient resolution and missing boundary information by combining CLIP's internal features with multi-resolution inputs, without relying on external models, thereby maintaining computational efficiency.


\section{MARS-CLIP}
\label{sec:proposed}

In this paper, we propose {\it MARS-CLIP}, a zero-shot image segmentation method designed to improve pixel-level dense prediction accuracy while maintaining the high generalization capability of CLIP.
Fig. \ref{fig:overview} illustrates the overview of our proposed framework.
MARS-CLIP consists of three steps.
The first step is feature extraction by multi-resolution inputs, where high-resolution feature maps integrating local details and global context are generated by partitioning the input image.
The second step involves refining the attention mechanism using low-level features.
We restore object boundary consistency by injecting spatial information from intermediate layers and color information as biases into the final attention layer.
The final step generates segmentation masks by calculating the similarity between the obtained image features and text features.


\subsection{Feature Extraction with Multi-resolution Images}
\label{ssec:multi}

Since the CLIP image encoder $E_{img}$ typically has a fixed input size, e.g., $224 \times 224$ pixels, directly feeding high-resolution images results in the loss of fine structural details due to downsampling.
To address this issue, we adopt a multi-resolution strategy that integrates local fine-grained features with global context features.
Specifically, given an input image $\mathbf{I} \in \mathbb{R}^{3 \times H \times W}$, we first apply reflective padding to generate $\mathbf{I}_{pad}$ such that its dimensions are multiples of the crop size $S_{crop}$.
Next, $\mathbf{I}_{pad}$ is divided into an $N \times M$ grid of local regions, and each region is resized to the encoder's input resolution $K \times K$.
Since typically $S_{crop} < K$, this process corresponds to zooming into local regions, enabling the extraction of fine-grained features.
The feature maps $\mathbf{F}_{local}^{(n,m)} = {E}_{img}(\mathbf{x}_{n,m})$ extracted from the image patch $\mathbf{x}_{n,m}$ at grid position $(n, m)$ are recombined according to their original spatial arrangement to form a single high-resolution feature map $\mathbf{F}_{local}$.
However, since processing local regions independently may result in the loss of global context, we also resize the entire image $\mathbf{I}$ and feed it into the encoder to extract a global feature map $\mathbf{F}_{global}$.
The final refined feature map $\mathbf{F}_{refined}$ is obtained by integrating these features as follows:
\begin{equation}
  \mathbf{F}_{refined} = \alpha \cdot \text{Up}(\mathbf{F}_{global}) + (1 - \alpha) \cdot \mathbf{F}_{local},
\end{equation}
where $\text{Up}(\cdot)$ denotes bilinear upsampling to align spatial resolutions, and $\alpha$ is a hyperparameter that balances the contributions of global and local information.

\begin{figure}[t]
  \centering
  \includegraphics[width=\linewidth]{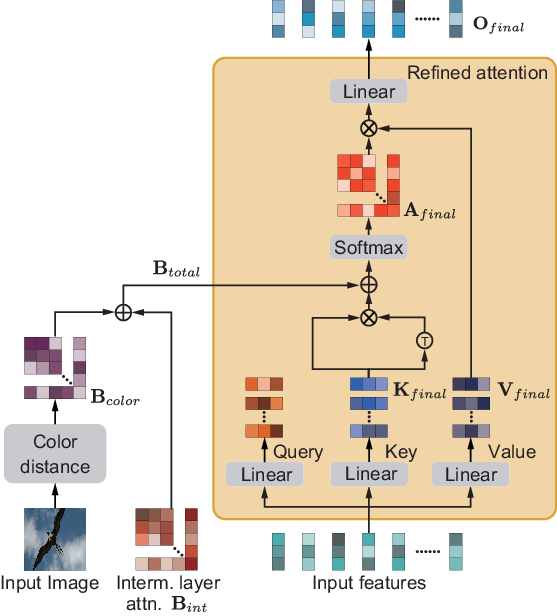} 
  \caption{Details of the proposed refined attention block.
  Unlike standard self-attention, we employ $\mathbf{KK}^\top$ similarity instead of $\mathbf{QK}^\top$ and inject spatial biases from intermediate layers along with color affinity biases from the input image.} 
  \label{fig:last_transformer} 
\end{figure}

\subsection{Structure-Aware Attention Refinement}
\label{ssec:atten_refine}

While the self-attention maps in the final layer of CLIP's ViT encoder possess strong capabilities for semantic grouping, they tend to exhibit spatially ambiguous object boundaries.
Recent methods, such as SCLIP \cite{Wang-ECCV-2024}, suggest that employing $\mathbf{KK}^\top$ instead of standard $\mathbf{QK}^\top$ attention improves segmentation performance.
However, boundary inconsistencies persist in complex scenes.
To address this issue, we inject spatial correlations from intermediate attention maps and color affinities derived from the input image as biases into the final attention mechanism, as shown in Fig. \ref{fig:last_transformer}.
The final attention map $\mathbf{A}_{final}$ is computed by adding a bias term $\mathbf{B}_{total}$ to the $\mathbf{KK}^\top$ similarity matrix followed by the Softmax function:
\begin{equation}
  \mathbf{A}_{final} = \text{Softmax}\left( \frac{\mathbf{K}_{final} {\mathbf{K}_{final}}^\top}{\sqrt{d}} + \mathbf{B}_{total} \right),
\end{equation}
where $\mathbf{K}_{final}$ denotes the Key features of the final layer, and $d$ is the feature dimension.
We compute the product of this attention map $\mathbf{A}_{final}$ and the final Value features $\mathbf{V}_{final}$, then apply a linear layer to obtain the output $\mathbf{O}_{final}$ of the refined attention block.
The bias term $\mathbf{B}_{total}$ is defined as the sum of an internal bias $\mathbf{B}_{int}$ based on intermediate attention maps and an external bias $\mathbf{B}_{color}$ based on color information.
For the internal bias $\mathbf{B}_{int}$, we utilize attention maps from specific intermediate layers that strongly preserve the spatial layout of the input image.
Furthermore, we integrate color information, which lacks semantic discriminability but retains precise boundary details, as the external bias $\mathbf{B}_{color}$.
Specifically, we convert the input image from sRGB to the CIELAB color space and use color features $\mathbf{c}_i \in \mathbb{R}^3$ downsampled to the patch level.
The color affinity between patches $i$ and $j$ is defined using a Gaussian kernel as follows:
\begin{equation}
  \mathbf{B}_{color}(i, j) = \exp \left( - \frac{\| \mathbf{c}_i - \mathbf{c}_j \|_2^2}{2\sigma^2} \right),
\end{equation}
where $\sigma$ is a bandwidth parameter, set to $\sigma = 30.0$ in our method.
Note that color bias is not applied to the class token since it lacks spatial positional information.
Additionally, following the insights from ClearCLIP \cite{Lan-ECCV-2024-Clear}, we remove the residual connections and the feed-forward network from the final Transformer block.
This prevents the original features from being dominated by residual components, thereby maximizing the effect of the refined attention mechanism.

\subsection{Zero-Shot Semantic Segmentation}

The final segmentation is performed by matching the obtained image features with the text features.
To generate text features, we insert the class name $y_k$ of the target dataset into the 80 predefined ImageNet templates \cite{Radford-ICML-2021}.
The text feature vector $\mathbf{F}_{text}^k$ is computed as the average of the embeddings obtained from the text encoder.
Regarding image features, let $\mathbf{F}_{refined}$ denote the high-resolution and boundary-aligned feature map.
We calculate the cosine similarity score $S_{i,j,k}$ between the image feature vector $\mathbf{F}_{refined}^{(i,j)}$ at pixel position $(i, j)$ and the text feature vector $\mathbf{F}_{text}^k$ for class $k$ as follows:
\begin{equation}
  S_{i,j,k} = \frac{\mathbf{F}_{refined}^{(i,j)} \cdot \mathbf{F}_{text}^k}{\| \mathbf{F}_{refined}^{(i,j)} \| \| \mathbf{F}_{text}^k \|}.
\end{equation}
Finally, the resulting similarity map is upsampled to the original image size, and the final prediction mask is obtained by assigning the class with the maximum score to each pixel.


\section{Experiments and Discussion}
\label{sec:experiments}

In this section, to verify the effectiveness of the proposed MARS-CLIP, we present ablation studies analyzing the contribution of each component and comparative experiments against state-of-the-art methods using standard benchmark datasets.


\subsection{Dataset and Evaluation Metrics}

We conduct evaluations on six standard segmentation benchmarks, following the protocols of previous studies such as SCLIP \cite{Wang-ECCV-2024} and NACLIP \cite{Hajimiri-WACV-2025}.
For simplicity, we define the following abbreviations for each dataset setting.
For PASCAL VOC 2012 \cite{Everingham-IJCV-2015} (1,449 images), we denote the 21-class setting including the background as ``V21'' and the 20-class setting excluding the background as ``V20''.
Similarly, for PASCAL Context \cite{Mottaghi-CVPR-2014} (5,105 images), we denote the 60-class setting as ``PC60'' and the 59-class setting as ``PC59''.
Additionally, we denote ADE20K \cite{Zhou-IJCV-2019} (2,000 images) as ``ADE'', COCO-Stuff \cite{Caesar-CVPR-2018} (5,000 images) as ``C-Stf'', Cityscapes \cite{Cordts-CVPR-2016} (500 images) as ``City'', and COCO-Object \cite{Lin-ECCV-2014} (5,000 images) as ``C-Obj''.
For quantitative evaluation, we use the mean Intersection over Union (mIoU), the standard metric in semantic segmentation, calculated as the average IoU across all classes.


\subsection{Implementation Details}

We use the pre-trained CLIP ViT-B/16 \cite{Radford-ICML-2021} as the backbone model and freeze the weights of both image and text encoders.
We employ 80 standard prompt templates for text embedding.
During inference, input images are resized such that the shorter side is 336 pixels (560 pixels for ``City'').
We then apply sliding window processing with a window size of $224 \times 224$ pixels and a stride of 112.
Hyperparameters specific to our method were set as follows based on the results of the ablation study.
The crop size for multi-resolution feature extraction is set to $112$, and the weight $\alpha$ for integrating global and local features is set to $0.8$.
For attention mechanism refinement, we utilize the intermediate attention maps from the 8th layer, which preserve spatial structure.
For final mask generation, Pixel Adaptive Mask Refinement (PAMR) \cite{Araslanov-CVPR-2020} is applied as post-processing, with the background class threshold set to $0.1$.
All experiments were implemented using PyTorch and conducted on an NVIDIA A100 GPU.


\begin{figure}[t]
  \centering
  \includegraphics[width=\linewidth]{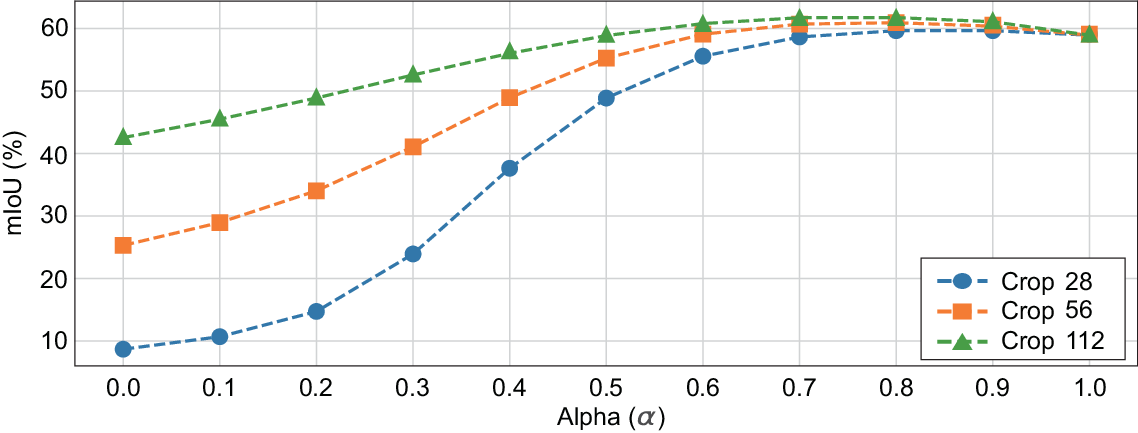} 
  \caption{Impact of fusion weight $\alpha$ across different crop sizes on ``V21''.} 
  \label{fig:weight_ablation} 
\end{figure}

\begin{table}[t]
  \centering
  \caption{Ablation study on attention refinement using intermediate features from the 8th layer.
  ``Feat. Sim.'' indicates cosine similarity between features.
  The bottom row (Attn. + Color) represents our proposed setting, which adds color similarity bias to the standard attention.}
  \label{tab:ablation_source_layer8}
  \setlength{\tabcolsep}{0pt} 
  \begin{tabular*}{\linewidth}{@{\extracolsep{\fill}}lcccccccc}
    \toprule
    Source & V21 & PC60 & C-Obj & V20 & City & PC59 & ADE & C-Stf \\
    \midrule
    Attn. ($\mathbf{QK}^\top$) & 58.6 & 32.6 & 33.8 & 83.0 & 32.7 & 35.9 & 17.1 & 23.9 \\
    Feat. Sim. & 58.0 & 32.4 & 33.6 & 82.6 & 32.0 & 35.6 & 16.9 & 23.8 \\
    $\mathbf{KK}^\top$ & 58.7 & 32.7 & 33.9 & 82.9 & 33.0 & 35.9 & 17.3 & 24.0 \\
    $\mathbf{QQ}^\top$ & 58.6 & 32.6 & 33.8 & 82.9 & 32.7 & 35.8 & 17.2 & 23.9 \\
    $\mathbf{VV}^\top$ & 58.5 & 32.6 & 33.8 & 82.8 & 32.7 & 35.8 & 17.1 & 23.9 \\
    \midrule
    \textbf{Attn. + Color} & \textbf{59.4} & \textbf{32.9} & \textbf{34.3} & \textbf{83.4} & \textbf{33.2} & \textbf{36.2} & \textbf{17.5} & \textbf{24.3} \\
    \bottomrule
  \end{tabular*}
\end{table}
\begin{table}[t]
  \centering
  \caption{Ablation study on the layer depth for extracting attention maps.
  ``Base'' denotes using only the intermediate attention map as a bias, while ``+Col'' indicates the addition of color similarity as a bias.}
  \label{tab:layer_wise_ablation}
  \small
  \setlength{\tabcolsep}{1.2pt}
  \begin{widetable}{\linewidth}{lcccccccccccc}
    \toprule
    Data & Meth. & L1 & L2 & L3 & L4 & L5 & L6 & L7 & L8 & L9 & L10 & L11 \\
    \midrule
    \multirow{2}{*}{V21} 
      & Base & 57.8 & 58.3 & 58.3 & 58.4 & 58.4 & 58.4 & \textbf{58.6} & \textbf{58.6} & 58.5 & 58.3 & 58.2 \\
      & +Col & 58.7 & 59.0 & 59.1 & 59.1 & 59.2 & 59.2 & 59.3 & \textbf{59.4} & 59.3 & 59.0 & 59.0 \\
    \midrule
    \multirow{2}{*}{PC60} 
      & Base & 32.4 & 32.5 & 32.5 & 32.5 & 32.5 & 32.5 & \textbf{32.6} & \textbf{32.6} & 32.5 & 32.4 & 32.4 \\
      & +Col & 32.8 & \textbf{32.9} & \textbf{32.9} & \textbf{32.9} & \textbf{32.9} & \textbf{32.9} & \textbf{32.9} & \textbf{32.9} & \textbf{32.9} & 32.8 & 32.8 \\
    \midrule
    \multirow{2}{*}{C-Obj} 
      & Base & 33.5 & 33.7 & 33.7 & 33.7 & \textbf{33.8} & \textbf{33.8} & \textbf{33.8} & \textbf{33.8} & \textbf{33.8} & 33.7 & 33.6 \\
      & +Col & 34.0 & 34.2 & 34.2 & 34.1 & 34.2 & 34.2 & \textbf{34.3} & \textbf{34.3} & 34.2 & 34.1 & 34.1 \\
    \midrule
    \multirow{2}{*}{V20} 
      & Base & 82.5 & 82.7 & 82.9 & 82.8 & 82.9 & 82.9 & 83.0 & 83.0 & \textbf{83.1} & 83.0 & 82.8 \\
      & +Col & 82.9 & 83.1 & 83.3 & 83.2 & 83.2 & 83.3 & \textbf{83.4} & \textbf{83.4} & \textbf{83.4} & 83.3 & 83.2 \\
    \midrule
    \multirow{2}{*}{City} 
      & Base & 31.8 & 32.2 & 32.5 & 32.4 & 32.6 & 32.6 & \textbf{32.7} & \textbf{32.7} & 32.5 & 32.2 & 32.2 \\
      & +Col & 32.4 & 32.7 & 33.0 & 32.9 & 33.1 & 33.1 & \textbf{33.2} & \textbf{33.2} & 33.0 & 32.8 & 32.7 \\
    \midrule
    \multirow{2}{*}{PC59} 
      & Base & 35.6 & 35.8 & 35.8 & 35.8 & 35.8 & 35.8 & 35.8 & \textbf{35.9} & 35.8 & 35.7 & 35.6 \\
      & +Col & 36.0 & 36.1 & 36.1 & 36.1 & 36.1 & 36.1 & \textbf{36.2} & \textbf{36.2} & 36.1 & 36.0 & 36.0 \\
    \midrule
    \multirow{2}{*}{ADE} 
      & Base & 16.9 & 17.0 & \textbf{17.1} & 17.0 & \textbf{17.1} & \textbf{17.1} & \textbf{17.1} & \textbf{17.1} & 17.0 & 17.0 & 17.0 \\
      & +Col & 17.3 & 17.4 & 17.4 & 17.4 & 17.4 & 17.4 & \textbf{17.5} & \textbf{17.5} & 17.4 & 17.3 & 17.3 \\
    \midrule
    \multirow{2}{*}{C-Stf} 
      & Base & 23.7 & \textbf{23.9} & \textbf{23.9} & \textbf{23.9} & \textbf{23.9} & \textbf{23.9} & \textbf{23.9} & \textbf{23.9} & \textbf{23.9} & 23.8 & 23.8 \\
      & +Col & 24.1 & 24.2 & 24.2 & 24.2 & 24.2 & 24.2 & 24.2 & \textbf{24.3} & 24.2 & 24.1 & 24.1 \\
    \bottomrule
  \end{widetable}
\end{table}
\begin{table}[t]
  \centering
  \caption{Ablation study on the components of MARS-CLIP.
  ``Multi'' refers to the adoption of the multi-resolution strategy, and ``Bias'' denotes attention refinement using low-level features.}
  \label{tab:ablation_selected}
  \small
  \begin{tabular*}{\linewidth}{@{\extracolsep{\fill}}ccccccc}
    \toprule
    Multi & Bias & V21 & City & ADE & C-Stf & Avg Gain \\
    \midrule
    --- & --- & 58.9 & 35.0 & 17.4 & 23.4 & --- \\
    --- & \checkmark & 59.9 & 35.5 & 17.5 & 23.4 & $+0.4$ \\
    \checkmark & --- & 61.7 & 37.7 & \textbf{18.5} & \textbf{24.5} & $+1.9$ \\
    \checkmark & \checkmark & \textbf{62.0} & \textbf{38.2} & \textbf{18.5} & 24.4 & $\bm{+2.1}$ \\
    \bottomrule
  \end{tabular*}
\end{table}
\begin{table*}[t]
  \centering
  \caption{Comparison with state-of-the-art methods.
  The ``PAMR'' column indicates the application of post-processing.
  Best results are \textbf{bold}.}
  \label{tab:result_main}
  \small
  \begin{tabular*}{\textwidth}{@{\extracolsep{\fill}}lccccccccc}
    \toprule
    Method & PAMR \cite{Araslanov-CVPR-2020} & V21 & PC60 & C-Obj & V20 & City & PC59 & ADE & C-Stf \\
    \midrule
    CLIP \cite{Radford-ICML-2021} & & 18.6 & 7.8 & 6.5 & 49.1 & 6.7 & 11.2 & 3.2 & 5.7 \\ 
    MaskCLIP \cite{Zhou-ECCV-2022} & & 43.4 & 23.2 & 20.6 & 74.9 & 24.9 & 26.4 & 11.9 & 16.7 \\
    CLIP Surgery \cite{Li-PR-2025} & & 41.2 & 30.5 & --- & --- & 31.4 & --- & 12.9 & 21.9 \\
    GEM \cite{Bousselham-CVPR-2024} & & 46.2 & --- & --- & --- & --- & 32.6 & 15.7 & --- \\
    SCLIP \cite{Wang-ECCV-2024} & & 59.1 & 30.4 & 30.5 & 80.4 & 32.2 & 34.2 & 16.1 & 22.4 \\
    ClearCLIP \cite{Lan-ECCV-2024-Clear} & & 51.8 & 32.6 & 33.0 & 80.9 & 30.0 & 35.9 & 16.7 & 23.9 \\
    NACLIP \cite{Hajimiri-WACV-2025} & & 58.9 & 32.2 & 33.2 & 79.7 & 35.5 & 35.2 & 17.4 & 23.3 \\
    \textbf{MARS-CLIP (Ours)} & & \textbf{62.0} & \textbf{33.8} & \textbf{34.5} & \textbf{81.1} & \textbf{38.2} & \textbf{36.8} & \textbf{18.5} & \textbf{24.4} \\
    \midrule
    SCLIP \cite{Wang-ECCV-2024} & \checkmark & 61.7 & 31.5 & 32.1 & 83.5 & 34.1 & 36.1 & 17.8 & 23.9\\
    NACLIP \cite{Hajimiri-WACV-2025} & \checkmark & 64.1 & 35.0 & 36.2 & 83.0 & 38.3 & 38.4 & 19.1 & 25.7\\
    \textbf{MARS-CLIP (Ours)} & \checkmark & \textbf{65.8} & \textbf{35.9} & \textbf{36.7} & \textbf{83.7} & \textbf{40.1} & \textbf{39.2} & \textbf{19.8} & \textbf{26.1} \\
    \bottomrule
  \end{tabular*}
\end{table*}
\begin{figure*}[t]
  \centering
  \includegraphics[width=.9\linewidth]{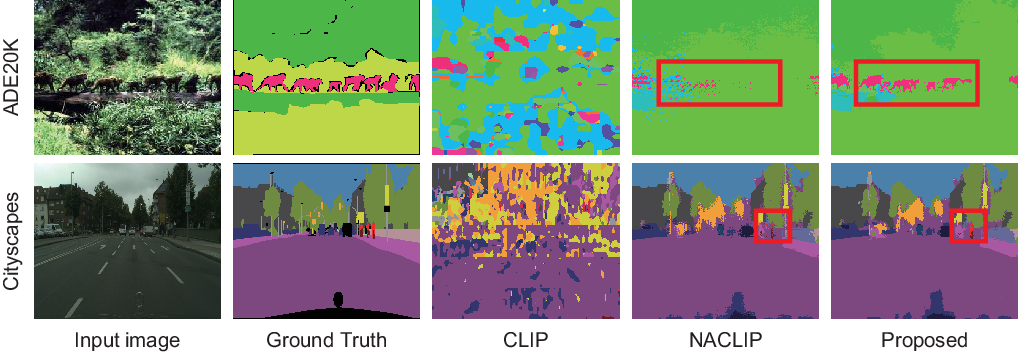}
  \caption{Qualitative comparison between existing methods and the proposed method.}
  \label{fig:result} 
\end{figure*}

\subsection{Ablation Study}

In this section, we validate the design choices of each component in MARS-CLIP and determine the optimal hyperparameters.
Specifically, we analyze parameter sensitivity in the multi-resolution strategy, the selection of feature sources for attention refinement, and the individual contributions of each proposed module.

First, we present the results of examining the integration weight $\alpha$ for global and local features and the crop size in multi-resolution feature extraction in Fig. \ref{fig:weight_ablation}.
The graph indicates that mIoU improves significantly when integrating both features compared to using single resolutions, such as $\alpha=0.0$ (local only) or $\alpha=1.0$ (global only).
In particular, performance tends to saturate and maximize around $\alpha=0.8$.
This suggests the importance of a balance where CLIP's inherent global semantic understanding serves as the backbone, while missing detailed information is moderately supplemented by local features.
Regarding crop size, 112 proved to be optimal.
Smaller crop sizes result in insufficient context within patches, leading to reduced recognition accuracy.
Based on these results, we fix the crop size to 112 and $\alpha$ to 0.8 for subsequent experiments.

Next, we investigate the optimal feature sources and extraction layers for refining the attention mechanism.
While our method uses intermediate attention maps and color information as biases, Table \ref{tab:ablation_source_layer8} presents a comparison with other features (self-correlations of $\mathbf{Q}$, $\mathbf{K}$, $\mathbf{V}$, and raw features).
The comparison reveals that the setting using attention maps $\mathbf{QK}^\top$ (Attn.) combined with color information (+Color) achieves the highest accuracy.
In particular, the gain from adding color information is substantial, confirming that low-level visual information contributes to defining object boundaries.
Table \ref{tab:layer_wise_ablation} shows the comparison regarding the layer depth for feature extraction.
Scores peak around the 8th layer (L8) and tend to decrease thereafter.
This can be interpreted as the loss of spatial layout information as feature abstraction progresses in deeper layers.
In our experiments, we adopt information from the 8th layer (L8), which showed the most stable performance on average.

Finally, Table \ref{tab:ablation_selected} presents the contribution of each component of our method.
Compared to the standard CLIP baseline, introducing the multi-resolution strategy (Multi) resulted in significant accuracy improvements.
Furthermore, combining this with attention mechanism refinement (Bias) led to further improvements, particularly on datasets containing high-definition and complex scenes such as ``City'' and ``ADE''.
These results demonstrate the complementary effects of enhancing resolution and restoring boundary information.





\subsection{Comparison with State-of-the-Art Methods}

We present the quantitative evaluation results on each benchmark in Table \ref{tab:result_main}.
To ensure a fair comparison, we conducted experiments under two settings: without post-processing and with post-processing  using PAMR \cite{Araslanov-CVPR-2020}.
Our proposed method, MARS-CLIP, achieved accuracy surpassing existing state-of-the-art methods across all datasets and settings.
In particular, our method demonstrates substantial performance improvement on datasets containing high-resolution images and fine-grained objects, such as ``City'' and ``ADE''.
For instance, without PAMR, MARS-CLIP recorded an mIoU of 38.2\% on ``City'', outperforming the runner-up NACLIP (35.5\%) by 2.7 points.
This suggests that the extraction of detailed features by the multi-resolution strategy and boundary refinement by color information bias effectively overcome the loss of fine structures caused by insufficient resolution, which is a common weakness in conventional methods.
Furthermore, the superiority of our method remains consistent even when PAMR is applied.
On ``City'', our method achieved 40.1\% mIoU, realizing a 1.8-point improvement over NACLIP \cite{Hajimiri-WACV-2025} (38.3\%).
This corroborates that the raw segmentation masks generated by our method already possess high spatial consistency without relying heavily on post-processing.
Qualitative evaluation results are shown in Fig. \ref{fig:result}.
Compared to the conventional method NACLIP \cite{Hajimiri-WACV-2025}, our method detects small objects and regions with complex boundaries more accurately.
These results demonstrate that the integrated mechanism of multi-resolution input and attention refinement in MARS-CLIP dramatically enhances the capability to recognize fine details in zero-shot segmentation.

We also compare inference cost on ``V21'' (shorter side 336 px, no sliding window).
MARS-CLIP requires 1.1 GB peak memory and 0.0403 s per image, compared to 1.1 GB / 0.0390 s for NACLIP \cite{Hajimiri-WACV-2025}.
The peak memory matches because the memory footprint is dominated by dense image--text similarity rather than multi-resolution processing, and the 0.0013 s latency overhead is negligible.

\section{Limitations}
The color affinity bias may degrade in low-light or low-contrast scenes.
Although the intermediate-layer spatial bias partially mitigates this, full illumination robustness is left as future work.

\section{Conclusion}
\label{sec:conclusion}

In this paper, we proposed MARS-CLIP, a zero-shot segmentation framework designed to accurately capture fine image details through multi-resolution inputs and attention mechanism refinement.
Our approach successfully restored object boundaries while preserving semantic consistency by achieving high-resolution feature representation through the integration of local and global features, and by injecting spatial information from intermediate layers and color information as biases into the final layer.
Through a set of experiments on six public datasets, we demonstrated the effectiveness of our proposed method, showing that it significantly outperforms state-of-the-art methods.


\section{Acknowledgment}
 
This work was supported in part by JSPS KAKENHI 23H00463 and 25K03131, JST BOOST JPMJBS2421, and the WISE Program for AI Electronics, Tohoku University.

{\small
  \bibliographystyle{IEEEbib}
  \bibliography{strings,refs}
}

\end{document}